\documentclass[10pt,twoside]{article}
\usepackage{float}
\usepackage{times}
\usepackage[utf8x]{inputenc}
\usepackage[frenchb]{babel}
\usepackage[T1]{fontenc}
\usepackage{graphicx}
\usepackage[colorlinks=true, allcolors=blue]{hyperref}
\usepackage{booktabs}
\usepackage{multirow}

\usepackage{taln2019}

\title{Qwant Research @DEFT 2019 : appariement de documents et extraction d'informations à partir de cas cliniques}

\author{Estelle Maudet, Oralie Cattan, Maureen de Seyssel,  Christophe Servan\\
  {\small
    QWANT RESEARCH, 7 Rue Spontini, 75116 Paris, France\\ 
    \texttt{
      initiale.nom@qwant.com\\ 
}}}

\begin{document}
\maketitle

\resume{
  Dans ce papier, nous présentons la participation de Qwant Research aux tâches 2 et 3 de l'édition 2019 du défi fouille de textes (DEFT) portant sur l'analyse de documents cliniques rédigés en français. La tâche 2 est une tâche de similarité sémantique qui demande d'apparier cas cliniques et discussions médicales. Pour résoudre cette tâche, nous proposons une approche reposant sur des modèles de langue et évaluons l'impact de différents pré-traitements et de différentes techniques d'appariement sur les résultats. Pour la tâche 3, nous avons développé un système d'extraction d'information qui produit des résultats encourageants en termes de précision. Nous avons expérimenté deux approches différentes, l'une se fondant exclusivement sur l'utilisation de réseaux de neurones pour traiter la tâche, l'autre reposant sur l'exploitation des informations linguistiques issues d'une analyse syntaxique.
}

\abstract{Document matching and information retrieval using clinical cases.}{
  This paper reports on Qwant Research contribution to tasks 2 and 3 of the DEFT 2019's challenge, focusing on French clinical cases analysis. Task 2 is a task on semantic similarity between clinical cases and discussions. 
  For this task, we propose an approach based on language models and evaluate the impact on the results of different preprocessings and matching techniques.
  For task 3, we have developed an information extraction system yielding very encouraging results accuracy-wise. We have experimented two different approaches, one based on the exclusive use of neural networks, the other based on a linguistic analysis.
}

\motsClefs
  {Similarité sémantique, extraction d'information, modèle de langues, modèle de vraisemblance de la requête, réseaux de neurones, analyse syntaxique}
  {Semantic similarity, information extraction, language model, query likelihood model, neural network, syntactic analysis}

\vspace{-0.25cm}\section{Introduction}

L'analyse et l'extraction d'informations pertinentes au sein d'un corpus médical est une tâche qui peut se montrer particulièrement difficile en raison de l'extrême spécificité du domaine. 
L'édition 2019 du défi fouille de texte (DEFT) porte sur cette problématique \cite{grabar2019DEFT}, et met à disposition un corpus de cas cliniques français, eux-mêmes issus du corpus CAS \cite{grabar2018cas}.

Nos motivations pour participer cette année au DEFT sont multiples. 
L'accent mis sur l'aspect médical de l'édition 2019 est particulièrement stimulant, du fait de l'impact médical d'éventuelles avancées dans le domaine. De plus, la spécificité du domaine, ses problématiques d'accès et la taille restreinte des ressources associées en font un défi particulièrement intéressant.
Il s'agit de notre première participation au défi. C'est pour nous une opportunité de nous confronter à d'autres équipes réfléchissant à des problématiques similaires.
Enfin, il nous tient à coeur de pouvoir contribuer à la recherche dans le domaine du traitement automatique des langues en France.

Nous avons participé à deux des trois tâches proposées dans le cadre de la campagne cette année. Nous détaillons dans la section \ref{sec:task2} notre contribution à la tâche 2 fondé sur l'appariement des cas cliniques et des discussions. La section \ref{sec:task3} décrit les méthodes employées dans le cadre de la tâche 3 dont le but est d'extraire des informations des cas cliniques.

\vspace{-0.25cm}\section{Tâche 2 : Mise en correspondance des cas cliniques et discussions par vraisemblance de la requête}
\label{sec:task2}

L'objectif de la tâche 2 est de faire un appariement entre un cas clinique et la discussion correspondante.
Le corpus d'entraînement contient 290 discussions et 290 cas cliniques. 
Tandis que chaque cas clinique est unique, plusieurs discussions peuvent être identiques.
Le corpus de test contient 214 discussions et cas cliniques, présentant les mêmes caractéristiques que le corpus d'entraînement.

\vspace{-0.15cm}\subsection{Approches}


L'approche utilisée dans le cadre de la tâche 2 est la même que celle proposée par \citet{Ponte1998LM} pour le calcul de la similarité basé sur des modèles de langue. 

L'idée principale est de générer un modèle de langue par discussion et de mesurer leur proximité avec chacun des cas cliniques, le plus proche étant celui qui sera apparié. 
La mesure utilisée est la perplexité, calculée suivant l'équation \ref{equa:ppl}:
\begin{equation}
    PPL=\hat{P}(w_1,...,w_m)^{-\frac{1}{m}}
    \label{equa:ppl}
\end{equation}

L'approche étant basée sur la forme de surface des mots, nous avons appliqué différents pré-traitements et étudié l'impact de ces derniers sur les résultats. Nous avons également étudié l'effet de différentes méthodes d'appariement de données. 



\vspace{-0.15cm}\subsection{Pré-traitement des données}
\label{sec:preprocessing}

Pour tenir compte des particularités linguistiques du domaine, nous avons appliqué un certain nombre de pré-traitements que nous détaillons dans cette partie.

Le texte est dans un premier temps converti en minuscules et tokenisé à l'aide de notre outil interne Qnlp-toolkit\footnote{https://github.com/QwantResearch/qnlp-toolkit}.
Différents autres traitements ont été appliqués en fonction des expérimentations: racinisation, suppression de mots vides et désabrègement.

\paragraph{Racinisation}
Aussi appelée dé-suffixation, la racinisation permet de regrouper l'ensemble des déclinaisons autour d'une même racine.
Elle a été réalisée suivant l'algorithme Snowball.  \cite{porter2001snowball}.
\paragraph{Suppression des mots vides}
Les mots très courants (souvent appelés \og~mots vides \fg), tels que \og~le \fg, \og~et \fg~  ou \og~de \fg, sont généralement ignorés dans les recherches. Ils ne contiennent habituellement pas autant d'information que les autres mots recherchés, qui eux permettent l'appariement des cas et discussions.
Une situation dans laquelle conserver les mots vides pourrait être important est la correspondance  de  l'information stylistique des cas et discussions (si les paires de cas et discussions avaient systématiquement été rédigées par la même personne).

\paragraph{Désabrègement}
Compte tenu de l'abondance  des sigles, acronymes, symboles et autres abrègements rencontrés dans les cas cliniques et dans les discussions, il nous a semblé intéressant de procéder à leur désabrègement automatique. 
Dans le domaine médical, l'abrègement est un procédé de construction lexicale très utilisé pour des raisons mnémotechniques et d'économie du langage qui permet de réduire les longues compositions de mots (souvent savants). Il arrive fréquemment que ces abrègements soient définis et repris tout au long du texte.
Afin de lever toute ambiguïté lexicale en écartant au maximum les cas d'abrègements polysémiques (e.g. IVG peut à la fois être utilisé pour signifier \og insuffisance ventriculaire gauche \fg~ou \og interruption volontaire de grossesse \fg), nous avons constitué un lexique des formes étendues, élaboré à partir des sigles, acronymes et de leurs définitions rencontrés en contextes à partir de l'ensemble du corpus. 
La mise en correspondance de la forme développée et de son abrègement a été réalisée selon plusieurs règles de recherche et nous avons complété notre lexique initial par des ressources terminologiques recensant les abrègements reconnus par la communauté. Ainsi, à partir du corpus comprenant les cas cliniques et les discussions s'y rapportant, 227 abrègements ont été relevés, ce qui a engendré un nombre de substitutions égal à 9016.

\vspace{-0.15cm}\subsection{Apprentissage des modèles de langue}

Les différents modèles de langue (LM) ont été appris à l'aide de l'outil \textit{SRILM} \cite{stolcke2002srilm}, selon trois ordres différents : unigrammes, bigrammes et trigrammes. 
Afin de permettre une comparaison entre les différentes perplexités (voir Section \ref{sec:task2_matchingtechniques}), 
tous les modèles ont été créés utilisant un vocabulaire commun, correspondant à l'ensemble des mots existants dans le corpus (cas cliniques et discussions).

\vspace{-0.15cm}\subsection{Appariements}
\label{sec:task2_matchingtechniques}


Les appariements peuvent se faire de différentes manières.
Nous avons choisi deux approches : apparier les cas cliniques avec les discussions et apparier les discussions aux cas cliniques (notées respectivement  \textit{c2d} et \textit{d2c}). 
Dans la première méthode, les modèles de langue sont entraînés sur les cas cliniques, et nous estimons la perplexité de chaque modèle sur les  discussions. 
Dans la seconde méthode, nous faisons l'inverse : nous entraînons les modèles de langue sur les discussions et nous estimons les scores de perplexité sur chacun des cas cliniques.

Nous avons également testé deux techniques permettant le choix des meilleures paires en fonction du score de perplexité. 
La première, non-exclusive (NE), consiste à choisir indépendemment et pour chaque LM le texte avec la perplexité la plus basse. 
Cela signifie que le même texte peut être apparié à plusieurs LMs. 
Dans la seconde technique, exclusive (E), un cas ne peut être apparié qu'une seule fois avec une discussion (et vice-versa). 
Pour ce faire, nous choisissons de façon itérative la paire de cas et discussion générant la perplexité la plus faible sur toutes les paires possibles, avant de supprimer ces cas et discussions de la liste de choix futurs. 
Cela exige que les scores de perplexités soient comparables, ce qui est le cas ici, le même vocabulaire ayant été utilisé pour générer tous les modèles de langue.

\vspace{-0.15cm}\subsection{Expériences \& Discussions}

Nous avons testé différentes combinaisons des techniques de pré-traitement et d'appariement présentées ci-dessus. 
Toutes les expériences ont été effectuées sur l'entièreté du corpus d'apprentissage, soit 290 paires de discussions et cas cliniques. 
Les scores obtenus sur le corpus d'évaluation pour les soumissions finales sont également présentés.

\vspace{-0.10cm}\subsubsection{Effet du pré-traitement de texte}

Les résultats présentés dans le tableau \ref{tab:task2-table-pretraitement} mettent en exergue l'effet de différentes techniques de pré-traitement sur les corpus de cas et discussions. 
Dans ce tableau, nous avons comparé uniquement les différentes approches en fonction de la méthode d'appariement \textit{d2c} décrite dans la section \ref{sec:task2_matchingtechniques}. De plus, tous les LMs sont d'ordre 2.
La racinisation est notée \textit{rac}, la suppression des mots-vides \textit{mv} et le désabrègement \textit{des}.

Le système initial, qui ne comporte aucun pré-traitement, atteint un score de 61,38 en précision et rappel.
Les pré-traitements classiques de racinisation et de suppression de mots-vides améliorent logiquement les scores de près de 11 points. 
le processus de désabrègement automatique, seul, offre des scores de précision et de rappel de 80,69, soit près de 19 points d'amélioration.
Lorsqu'on combine les deux pré-traitements, malheureusement, les améliorations ne se cumulent pas.
Au contraire, on observe une légère contre-performance de 0,7 points par rapport au meilleur système.

\begin{table}[h!]
\centering
\begin{tabular}{lccc}
\hline
Pré-traitement     & Apprentissage \\
\hline
Initial    &  61,38  \\
\hline
rac+mv     & 72,41  \\
des        & \textbf{80,69}   \\
rac+mv+des & 80,00  \\
\hline
\end{tabular}
\caption{Scores de précision sur les données d'apprentissage de la tâche 2, en fonction du pré-traitement testé. Tous les LMs sont d'ordre 2, et l'appariement s'est fait de façon exclusive, en direction d2c.
\textit{rac: racinisation; mv: mots vides supprimés; des: désabrègement.}}
\label{tab:task2-table-pretraitement}
\end{table}

\vspace{-0.10cm}\subsubsection{Effet de l'ordre du modèle de langue}

Utilisant des modèles de langue, nous nous sommes intéressés à l'impact de l'ordre de ces derniers.
Le tableau \ref{tab:ngram} présente les résultats obtenus.
Nous avons utilisé une configuration \textit{d2c}, avec le pré-traitement de désabrègement.
On peut constater que les modèles de langue d'ordre 2 et 3 obtiennent de meilleurs résultats que les modèles d'ordre 1 (amélioration de près de 3 points à l'ordre 2).

\begin{table}[h!]
 \centering
 \begin{tabular}{lc}
\hline
Ordre & Précision  \\
 \hline
 1-gram & 77,24       \\
 2-gram  &  \textbf{80,68}   \\
 3-gram & 79,31 \\ 
 \hline
 \end{tabular}
 \caption{Scores de précision pour les données de la tâche 2, en fonction de l'ordre du LM. L'appariement s'est fait de façon exclusive, en direction \textit{d2c}. Le corpus d'apprentissage a été pré-traité uniquement avec les abréviations normalisées (\textit{des}).}
 \label{tab:ngram}
 \end{table}

\vspace{-0.10cm}\subsubsection{Impact des techniques d'appariement} \label{impact-appariement}

Nous avons également testé les différentes techniques d'appariement introduites dans la Section \ref{sec:task2_matchingtechniques}. 
La Table \ref{tab:appariement} souligne l'amélioration apportée par la technique d'exclusivité, avec un score de précision et de rappel systématiquement plus haut que pour les mêmes systèmes n'utilisant pas cette technique. 
En effet, sans ce procédé, il est probable que si un cas est relativement général dans les termes qui le composent, il soit apparié à une grande majorité de discussions (ou vice-versa). 
Puisque la tâche 2 nécessite qu'un texte ne soit apparié qu'une seule fois, nous avons choisi d'utiliser cette technique dans nos soumissions.  

L'importance de la direction utilisée pour l'appariement (\textit{d2c} ou \textit{c2d} - voir Section \ref{sec:task2_matchingtechniques}), est aussi mise en exergue dans les résultats présentés Table \ref{tab:appariement}. 
Il semble ainsi que mesurer la proximité de modèles de langue estimés sur les discussions par rapport à chacun des cas cliniques (\textit{d2c}) produise des résultats plus probants que dans le cas inverse. Les résultats peuvent s'expliquer par la plus grande longueur des discussions par rapport aux cas (environ 393 mots en moyenne pour les cas versus 919 mots pour les discussions). Les discussions permettant ainsi d'estimer des modèles de langue plus variés. Une expérience intéressante pour le futur serait de combiner les deux approches, et sélectionner les meilleurs scores sur toutes les paires possibles, avec les deux directions \textit{c2d} et \textit{d2c}.

\begin{table}[h!]
 \centering
 \begin{tabular}{lcc}
\hline
Direction & NE/E &  Précision \\
 \hline
c2d & NE	 & 40,34  \\
c2d & 	E	 & 71,03 \\
d2c	 & NE & 	48,96  \\
d2c	 & E & 	\textbf{80,69}\\
 \hline
 \end{tabular}
 \caption{Scores de précision pour les données de la tâche 2, en fonction de la méthode d'appariement. L'appariement s'est fait de façon exclusive (\textit{E}) ou non-exclusive (\textit{NE}), dans les deux directions (\textit{d2c} et \textit{c2d}). Le corpus d'apprentissage a été pré-traité uniquement avec les abréviations normalisées (\textit{des}).}
 \label{tab:appariement}
 \end{table}



\vspace{-0.10cm}\subsubsection{Résultats soumis}

La Table \ref{tab:task2-results} récapitule les résultats obtenus sur les trois contributions que 
nous avons soumis pour la tâche 2 de DEFT 2019. En accord avec les conclusions tirées Section \ref{impact-appariement}, nous utilisons l'approche d2c et la technique d'exclusion pour les trois soumissions. 

La première version testée (\textit{run-1}) correspond aux meilleurs résultats obtenus lors de tous nos essais sur le corpus d'apprentissage. 
Le texte a été racinisé, désabrégé et les mots vides supprimés. 
Les modèles de langue sont d'ordre 1, et ont été créés sur les discussions (direction \textit{d2c}).  
Pour la deuxième soumission (\textit{run-2}), nous avons choisi l'approche qui serait théoriquement la meilleure en se basant exclusivement sur les conclusions tirées lors des expériences sur le corpus d'apprentissage. 
Nous avons ainsi utilisé un modèle de langue d'ordre 2 et le seul pré-traitement effectué est le désabrègement. 
Enfin, la troisième version (\textit{run-3}) est plus expérimentale.
Nous avons décidé de jouer sur l'ordre du modèle de langue (choisissant un modèle d'ordre 3) et d'utiliser outre cette variable les mêmes caractéristiques que pour la version 1 (désabrègement, racinisation, suppression des mots vides, direction \textit{d2c} et technique d'exclusion).


 \begin{table}[h]
 \centering
 \begin{tabular}{@{}lllll@{}}
 \toprule
   \multirow{2}{*}{Version} & \multirow{2}{*}{Ordre} & \multirow{2}{*}{Pré-traitement}     & \multicolumn{2}{c}{Précision} \\
 & & 
 & Apprentissage & \'Evaluation \\
 \midrule
 run-1 & 1-gram & rac+mv+des & \textbf{81,72}  & \textbf{84,11}   \\
run-2 & 2-gram & des & 80,69  & 76,17  \\
run-3 & 3-gram & rac+mv+des & 80,34  & 83,18 \\ \bottomrule
 \end{tabular}
 \caption{Scores de précision pour les résultats soumis pour la tâche 2.  L'appariement s'est systématiquement fait de façon exclusive, en direction \textit{d2c}.}
 \label{tab:task2-results}
 \end{table}

Les approches 1 et 3 produisent les résultats attendus sur les données de test, proches
de ceux obtenus sur le corpus d'entraînement. 
La seconde version (run-2) cependant, produit sur le corpus de test des résultats en deçà de ceux obtenus sur le corpus d'apprentissage. 
Puisque nous savons que l'impact de l'ordre des modèles de langue est restreint, il est probable que ces résultats viennent du pré-traitement choisi. 
Cela peut souligner la possible importance de la racinisation et de la suppression de mots-vides lors de l'utilisation de corpus de taille limitée.

Il est également intéressant d'observer que l'approche donnant lieu aux meilleurs résultats utilise des modèles de langue d'ordre 1. 
Ce type de modèle de langue (ou modèle \og sac de mots \fg), qui ne prend en compte que la fréquence des mots, ignorant leur ordre, peut donc suffire pour ce type d'exercice. 
En effet, la taille extrêmement restreinte des cas et discussions sur lesquelles les modèles de langue ont été estimés ne permet aucun gain d'information en utilisant des modèles d'ordre plus élevé.

\vspace{-0.25cm}\section{Tâche 3 : Extraction d'information sur des cas cliniques}
\label{sec:task3}

La tâche 3 est une tâche d'extraction d'information de type démographique et médicale. 
Ses  objectifs concernent  l'identification de cinq types d'informations correspondant au moment du dernier élément clinique rapporté dans le cas : l'âge et le genre de la personne concernée, l'origine (motif de la consultation ou de l'hospitalisation) et l'issue parmi cinq valeurs possibles (guérison, amélioration, stable, détérioration, décès). 
Pour tous ces cas, il est possible qu'une ou plusieurs des informations soient manquantes. 
Dans cette situation, la valeur est 'NUL'.

\vspace{-0.15cm}\subsection{Approches}

Nous présentons ci-dessous deux approches utilisées pour tenter de résoudre cette tâche: une approche neuronale et une approche hybride intégrant des connaissances linguistiques.
L'évaluation de leur pertinence sera ensuite proposée.

\vspace{-0.10cm}\subsubsection{Pré-traitements des données}

La tokénisation et la suppression de la casse ont été réalisées à l'aide de notre outil interne Qnlp-toolkit\footnote{https://github.com/QwantResearch/qnlp-toolkit}.

La lemmatisation permet d'obtenir la forme canonique des mots. 
Elle trouve son intérêt dans le cadre de ce travail car elle permet de débarrasser les mots des marques d'inflexion telles que celles de genre (masculin, féminin), de pluriel ou de conjugaison. La lemmatisation d'un verbe est la forme à l'infinitif de ce verbe, celle d'un nom, adjectif ou déterminant, sa forme au masculin singulier. Les mots, ou plus précisément les chaînes de caractères, peuvent ainsi être comparés à un niveau plus fin.  
Elle est effectuée en utilisant des règles de correspondance à partir des données de WordNet \cite{fellbaum1998wordnet}.

La lemmatisation a été appliquée uniquement dans le cadre de l'approche hybride. 
En effet, les modèles neuronaux se basent sur des représentations vectorielles des mots et nécessitent pas de pré-traitement si ce n'est la tokénisation. 

\vspace{-0.10cm}\subsubsection{Approche neuronale}

Nous avons utilisé un modèle neuronal supervisé pour l'étiquetage des empans correspondant aux informations à extraire des cas cliniques en adoptant un schéma d'étiquetage BIO \cite{ramshaw1999text}. 
L'approche choisie a été proposée dans \citet{ma2016end}. Elle a l'avantage de ne pas nécessiter un volume de données important pour l'apprentissage et obtient des performances au niveau de l'état de l'art pour la reconnaissances des entités nommées ou l'étiquetage en parties du discours.

\begin{figure}
    \centering
    \includegraphics[width=0.9\linewidth]{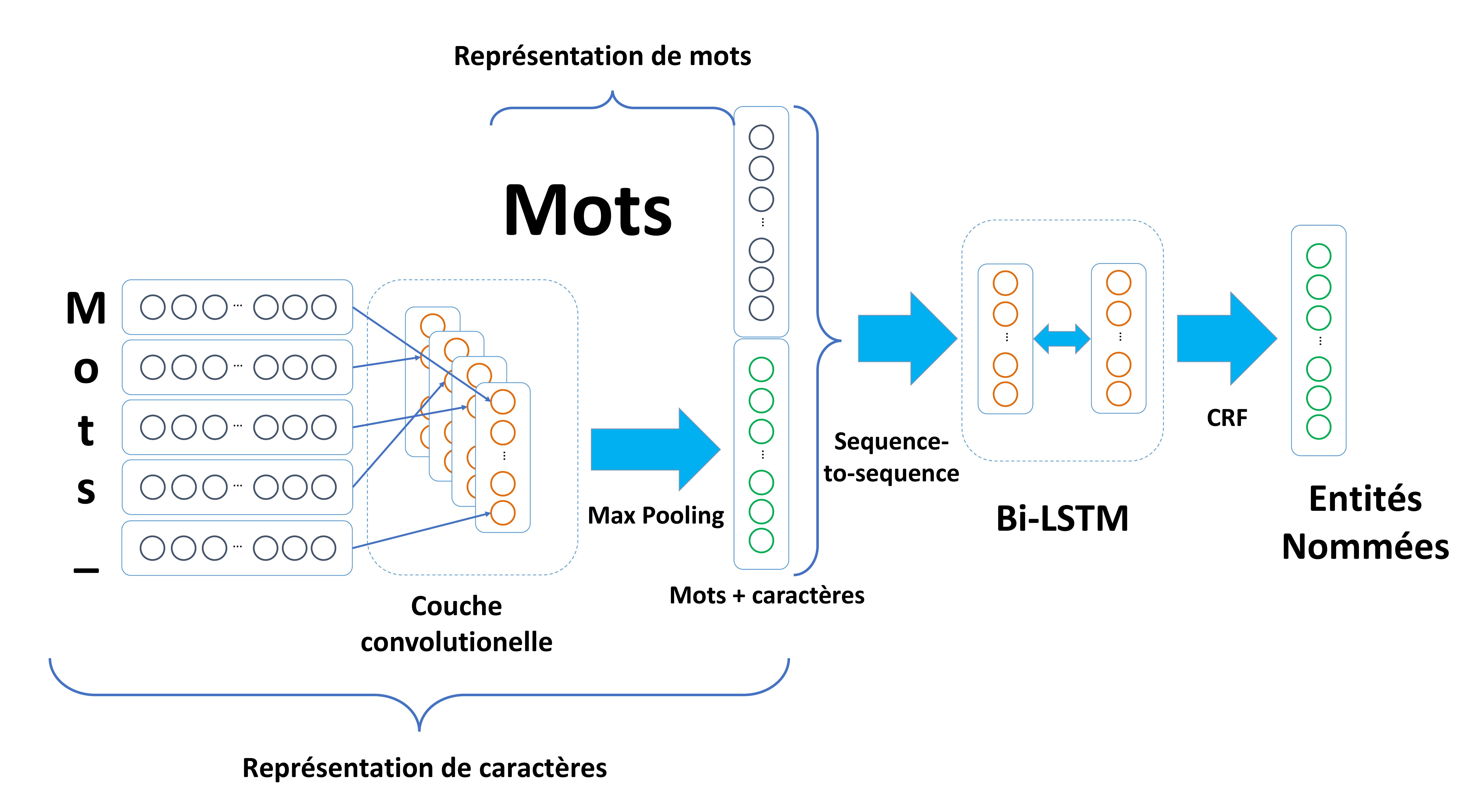}
    \caption{Modèle neuronal utilisé pour l'extraction d'entités nommées fondé sur l'approche proposée par \citet{ma2016end}.}
    \label{fig:NER}
\end{figure}

Les résultats sont rendus possibles grâce à l'utilisation de représentations pré-entraînées de mots et de caractères ainsi qu'à la combinaison d'un réseau de neurones récurrent (\textit{Bi-directional Long-Short Term Memory}, Bi-LSTM), un réseau de neurones convolutionnel (CNN) et un champ markovien conditionnel (CRF) tels que présentés dans la figure \ref{fig:NER}.

Pour créer le corpus d'apprentissage, nous avons manuellement étiqueté les 290 cas cliniques selon 4 types d'étiquettes : Âge, Genre, Origine de l'admission et Issue.

Dans le but d'améliorer les performances du modèle et pour palier au faible nombre de données, nous avons généré automatiquement des introductions de cas typiques. En effet, les premières lignes comportent très souvent trois des quatre informations à extraire: âge, genre et raison.
Cette génération fut opérée grâce à une grammaire hors-contexte ainsi qu'un certain nombre de ressources linguistiques, telles que des listes de symptômes et de maladies.
Les cas suivants ont par exemple pu être générés:
\begin{itemize}
    \item \textit{mlle a a été mis sous observations suite à hématome temporo-pariétal post-traumatique} ;
    \item \textit{mme u , 10 ans , a été traité pour une cancer épidermoïde} ;
    \item \textit{un jeune homme agé de 20 ans avec comme antécédent un perforation digestive instrumentale , a été mis sous observations suite à une surinfection kt jugulaire}.
\end{itemize}

Un total de 2000 cas ont été générés automatiquement et intégrés aux données d'apprentissage.

En parallèle, nous avons appris sur des corpus plus larges des représentations de mots afin de les utiliser dans le modèle. 
Nous avons extrait l'ensemble des pages du Wikipédia français appartenant au portail de la Médecine \footnote{Le portail médecine regroupe les articles appartenant au domaine médical \url{https://fr.wikipedia.org/wiki/Portail:M\%C3\%A9decine}.} ainsi que le corpus EMEA contenant des documents PDF de l'agence européenne de Médecine \footnote{Le corpus EMEA est accessible à l'adresse \url{http://opus.nlpl.eu/EMEA.php}.}. 
Sur ce corpus agrégé, nous avons appris des représentations de mots en utilisant l'outil FastText\footnote{FastText: \url{https://fasttext.cc}} proposé par \citet{bojanowski2017enriching}.

Après l'étiquetage du texte, nous inférons les informations demandées, notamment concernant l'âge, le genre, ainsi que l'issue.
En effet, une fois l'obtention d'un empan annoté comme "Âge", nous pouvons déduire l'âge en années.
A partir d'un certain nombre de règles heuristiques, nous inférons la valeur pour un empan tel que \textit{"18 mois"}. Le résultat obtenu est alors 1 an.
De la même manière, un certain nombre d'heuristiques ont été utilisées pour inférer le genre du patient à partir de l'empan correspondant.
Concernant l'origine de l'admission, aucune modification n'a été appliquée à l'empan retourné par le modèle.

L'identification de l'issue est quant à elle vue comme un problème de classification multi-classes où l'on considère ses cinq valeurs possibles (guérison, amélioration, stable, détérioration, décès) plus une sixième, utilisée pour nous permettre de considérer les cas où la valeur est NUL. 
Cette classification repose sur un modèle neuronal proposé par \citet{joulin2017bag} et est réalisée avec fastText.

\vspace{-0.10cm}\subsubsection{Approche hybride}

Une méthode alternative à la précédente a été explorée. Elle se fonde sur des connaissances linguistiques pour extraire les unités lexicales correspondant aux âges, genres et origines recherchés, à partir de l'analyse syntaxique en dépendance des cas cliniques et l'utilisation de patrons lexico-syntaxiques prédéfinis.

La  définition des  patrons se déroule selon plusieurs étape. Les textes sont dans un premier temps segmentés en phrases puis filtrés selon la pertinence des informations qui y sont présentes. Les phrases sélectionnées sont ensuite analysées syntaxiquement et les patrons sont construits à partir de leurs analyses.

\begin{figure}[ht]
    \centering
    \includegraphics[width=0.9\linewidth]{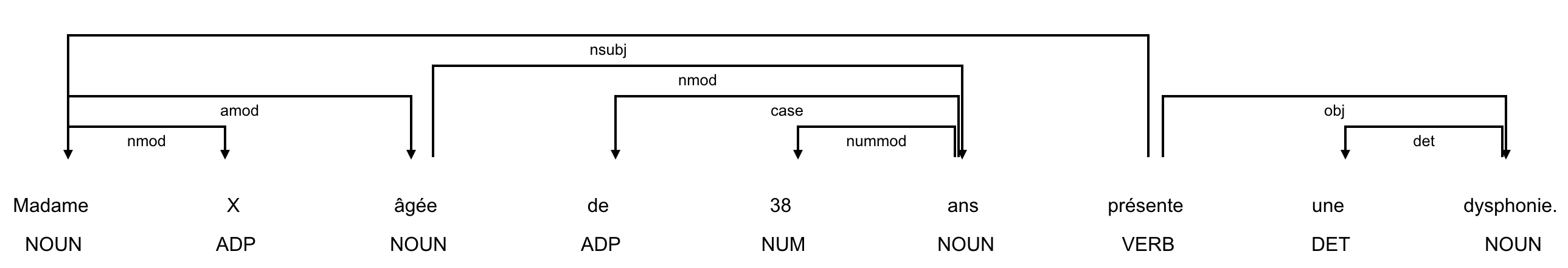}
    \caption{Exemple de phrase sélectionnée après analyse syntaxique.}
    \label{fig:syntax}
\end{figure}

Certains patrons sont plus productifs que d'autres. On observe par exemple qu'il existe 148 cas où l'âge est exprimé au moyen de l'expression "être âgé de" (Figure \ref{fig:syntax}). Dans 268 cas, il est extrait à partir du chemin de dépendance du modifieur nominal 'an'.
Ce dernier peut guider la recherche du genre, à partir du syntagme nominal sujet de l'arbre. L'identification du genre est alors réalisée en se référant aux valeurs des traits morpho-syntaxiques de ses constituants.

Dans quelques cas, l'âge n'est pas explicité dans le texte et doit être inféré à partir de dates, par exemple "[...] né le 25 avril 1972 , a consulté en 1996 [...]" (valeur attendue pour âge : 24). Il faut également pouvoir tenir compte des expressions nominales telles que «nouveau né», «quinquagénaire», etc. qui renseignent l'âge.

Pour l'origine, les représentations arborescentes extraites correspondant aux compléments verbaux (syntagmes nominaux et syntagmes prépositionnels) en position postverbale entretenant une relation directe avec les verbes identifiés nous a permis de repérer et d'extraire un ensemble de lexies verbales apparaissant de façon récurrente et se révélant toutes se référer à la tâche de prise en charge (présenter, hospitaliser, admettre, consulter, adresser, etc.). 

En ce qui concerne l'identification des issues, la version lemmatisée d'une liste de lexies spécifiques au champ sémantique lié au décès a permis d'identifier les cas de décès, ensuite un classifieur est entraîné pour identifier les autres cas.

\vspace{-0.15cm}\subsection{Expériences \& Discussions}

\vspace{-0.10cm}\subsubsection{Corpus d'apprentissage}

Dans  cette  section,  nous  mesurons la  capacité  de nos  méthodes à extraire les  informations recherchées.

\paragraph{Étiquetage automatique}
\label{sec:étiquetage}

Nous présentons tout d'abord les résultats de l'étiquetage automatique du texte.
Nous avons retranché dix pour-cent de cas pour l'évaluation. 
La table \ref{task3-ner-scores-train} présente les scores de F-measure pour chacun des types d'étiquette. 
Trois cas sont présentés, tout d'abord un apprentissage du modèle sur les données d'apprentissage uniquement, puis l'utilisation de représentations de mots pré-entraînées sur des corpus de données médicales (PRE). Enfin, nous évaluons l'ajout de données générées à l'aide d'une grammaire hors-contexte (GEN).

\begin{table}[h]
\centering
\begin{tabular}{llllll}
\toprule
Modèle / F-measure    & Age & Genre & Issue & Origine & All \\ \midrule
Étiqueteur & 84.21 & 50.00  & 51.72  & 28.57  & 53.81 \\
Étiqueteur + PRE & 87.72  & 59.26 & 47.27 & 36.73 & 58.60 \\ 
Étiqueteur + GEN + PRE & 90.00 & 53.33 & 58.62 & 43.64 & 61.80 \\ \bottomrule
\end{tabular}
\caption{Scores de F-measure sur l'étiquetage automatique de la tâche 3 pour un modèle appris sur 90\% du corpus d'entraînement et testé sur le reste. PRE désigne l'utilisation de représentations de mots pré-entraînées sur des corpus de données médicales. GEN correspond à l'extension des données d'apprentissage par des données générées automatiquement à l'aide d'une grammaire hors-contexte.}
\label{task3-ner-scores-train}
\end{table}

La combinaison la plus efficace est celle qui regroupe l'utilisation des représentations de mots pré-entraînées et les données générées automatiquement. 
Le nombre de données du corpus d'apprentissage étant relativement restreint, tout ajout de connaissances extérieures permet d'augmenter sensiblement les performances du modèle.

Nous décidons de conserver le modèle avec utilisation de représentations de mots pré-entraînées ainsi que les données générées automatiquement.

\paragraph{Classification de l'issue}

Les résultats de la classification de l'issue sont présentés dans la table~\ref{task3-issue-scores-train}. 
L'évaluation s'est faite par validation croisée avec 10 plis.

Le premier modèle évalué est appris sur l'ensemble du cas clinique. Étant donné que seule une partie du cas clinique se réfère directement à l'issue finale, nous avons étudié un modèle basé uniquement sur la fin du cas clinique, en se limitant aux dernières phrases.
De plus, nous considérons aussi des modèles appris uniquement sur les empans annotés comme "Issue". Nous évaluons, dans un premier temps, un modèle appris et testé sur les empans annotés manuellement pour rendre compte de la validité de l'approche. Dans un second temps, nous testons ce même modèle sur les empans annotés automatiquement.

\begin{table}[h!]
\centering
\begin{tabular}{lllll}
\toprule
Portion du cas clinique                            & Précision & Rappel & F-measure &  \\ \midrule
Cas clinique entier                                & 0.4482    & 0.4459 & 0.4471    &  \\
Deux dernières phrases du cas clinique             & 0.4448    & 0.4448 & 0.4448    &  \\ \midrule
Empan de l'issue obtenu par étiquetage manuel      & 0.6215    & 0.6215 & 0.6215    &  \\
Empan de l'issue obtenu par étiquetage automatique & 0.4222    & 0.4222 & 0.4222    &  \\ \bottomrule
\end{tabular}
\caption{Scores de précision, rappel et F-measure sur l'issue en validation croisée. On compare les résultats sur différentes portions de texte du cas clinique. D'un côté, on considère l'entièreté du cas clinique ainsi que les deux dernières phrases du cas clinique. De l'autre, on considère uniquement les empans étiquetés comme "Issue". Ces empans sont obtenus manuellement dans un cas et automatiquement dans l'autre.
}
\label{task3-issue-scores-train}
\end{table}

On observe que le meilleur score est celui obtenu en prédisant l'issue uniquement sur la partie du texte s'y référant.
Malheureusement, lorsque l'empan est obtenu par étiquetage automatique et non pas manuel, la qualité est grandement dégradée.
Cela est dû à la faible qualité de l'étiquetage automatique de l'issue présenté dans le paragraphe précédent.

Pour la phase de test, on décide de conserver le cas où l'on considère le document en entier ainsi que celui basé sur les empans sélectionnés seulement.

\vspace{-0.10cm}\subsubsection{Corpus de test}

Après avoir choisi les systèmes les plus prometteurs à partir des résultats obtenus sur le corpus d'entraînement, nous avons pu évaluer nos approches bout-à-bout sur le corpus de test.

\paragraph{Âge et genre}

Les résultats concernant l'âge et le genre sont présentés dans la table \ref{task3-age-gender-scores-test}.
L'approche neuronale d'étiquetage de mots et l'approche par arbre syntaxique sont comparées. 
On observe que l'approche par étiquetage neuronal fonctionne mieux que l'approche lexico-syntaxique. 
En effet, malgré le faible nombre de données à l'origine (290 données d'apprentissage), l'extension des cas cliniques par génération automatique ainsi que le recours aux représentations de mots pré-entraînées permettent d'obtenir une bonne généralisation, et donc des résultats satisfaisants pour l'étiquetage automatique.

\begin{table}[h!]
\centering
\begin{tabular}{l|ccc|ccc}
\toprule
\multirow{2}{*}{Approche}                  & \multicolumn{3}{c|}{Age} & \multicolumn{3}{c}{Genre}  \\
                  & Précision    & Rappel  & F1  & Précision     & Rappel   & F1   \\ \midrule
Étiqueteur + PRE + GEN (\textit{run-1})& \textbf{0.9748}       & \textbf{0.9023}  &  \textbf{0.9371} & 0.9421        & 0.9465 & 0.9442      \\
Analyse lexico-syntaxique & 0.9719       & 0.8860 & 0.9269    & \textbf{0.9555}        & \textbf{0.9488} & \textbf{0.9521}      \\ \bottomrule                                 
\end{tabular}
\caption{Scores de précision, rappel et F-measure (F1) pour l'extraction de l'âge et du genre sur le corpus de test. Étiqueteur + PRE + GEN correspond au système envoyé au soumission en première tentative (\textit{run-1}). 
L'analyse lexico-syntaxique utilise une approche par arbre syntaxique pour extraire les informations.}
\label{task3-age-gender-scores-test}
\end{table}

\paragraph{Origine de l'admission}
\begin{table}[h]
\centering
\begin{tabular}{l|ccc|ccc|c}
\toprule

\multirow{2}{*}{Approche} & & & & & & & \textit{micro} \\
                  & \multicolumn{3}{c|}{macro} & \multicolumn{3}{c|}{micro} & \textit{overlap} \\
                  & Pr    & Rp  & F1  & Pr & Rp & F1 & \textit{accuracy} \\ \midrule

Étiqueteur + PRE + GEN (\textit{run-1}) & 0.785 & 0.579 & 0.666 & 0.658 & 0.640 & 0.649 & 0.589 \\ \bottomrule
\end{tabular}
\caption{Scores de micro et macro précision, rappel, F-measure et \textit{micro overlap accuracy} pour l'extraction de l'origine de l'admission sur le corpus de test à partir d'un système étiqueteur neuronal avec représentations de mots pré-entraînées et génération de données d'apprentissage.}
\label{task3-admission-scores-test}
\end{table}
Les résultats relatifs à l'admission sont présentés dans la table \ref{task3-admission-scores-test}. 
Les scores issus de l'étiqueteur neuronal sont prometteurs et les différences entre précision et rappel (macro et micro) laissent supposer que le modèle retourne un empan de texte trop précis.
Nous n'avons pas pu obtenir de résultats concluants par analyse lexico-syntaxique car, contrairement à l'âge et au genre, les cas cliniques ne suivent pas un schéma suffisamment récurrent pour obtenir une bonne extraction.

\paragraph{Issue}
Les résultats de l'issue sont présentés dans la table \ref{task3-issue-scores-test}. 
Dans les deux premiers cas, nous avons utilisé uniquement un modèle de classification pour prédire l'ensemble des classes. 
Nous avons tout d'abord évalué un modèle appris à partir de l'entièreté du cas clinique. 
Nous avons ensuite comparé ses résultats avec ceux d'un autre classifieur entraîné à prédire sur les données étiquetées automatiquement. 
Enfin, nous avons identifié les cas de décès en utilisant des connaissances linguistiques puis appris un modèle pour les issues restantes sur les quatre dernières phrases des cas cliniques. 
Après annotation manuelle du corpus d'entraînement pour étiquetage, nous avons en effet fait plusieurs observations relatives à l'issue. 
D'une part, les cas de décès se prêtent mieux à une approche lexicale avec un vocabulaire très spécifique. 
Et d'autre part, dans la majorité des cas, l'empan de texte renvoyant à l'issue apparaît vers la fin du cas clinique. 
Cette dernière approche est celle qui retourne les meilleurs résultats avec un score de 0.60 et 0.58 en précision et rappel respectivement.

\begin{table}[h!]
\centering
\begin{tabular}{l|ccc}
\toprule
\multirow{2}{*}{Approche}                    & \multicolumn{3}{c}{Issue}\\
                  & Précision    & Rappel & F1  \\ \midrule
Cas clinique entier                                      & 0.5285 & 0.5199 & 0.5241 \\
Empan de l'issue obtenue par étiquetage (\textit{run-1}) & 0.5198 & 0.4918 & 0.5054   \\ \midrule
Traitement linguistique de \textit{décès} et quatre dernières phrases & \textbf{0.5985} & \textbf{0.5831} & \textbf{0.5906}   \\ \bottomrule
\end{tabular}
\caption{Scores de précision, rappel et F-mesure pour l'issue sur le jeu de test. Le première approche considère l'ensemble du cas clinique. Une seconde approche utilise uniquement l'empan "Issue" obtenu par étiquetage automatique. Enfin, une dernière approche utilise un traitement linguistique pour la détection de l'issue "décès", tandis que les quatre dernières phrases sont considérés pour prédire les autres issues.}
\label{task3-issue-scores-test}
\end{table}

\vspace{-0.25cm}\section{Conclusion}

Dans cet article, nous avons présenté notre participation aux tâches 2 et 3 proposées dans le cadre du DEFT 2019, correspondant respectivement aux tâches de recherche de similarité sémantique et d'extraction d'information, appliquées au domaine médical.

La méthode utilisée pour la tâche 2 repose sur l'utilisation de modèles de langue pour estimer la similarité entre documents en ne nous restreignant qu'aux données fournies. Elle a ensuite été étendue à des essais sur l'impact des pré-traitements utilisés en faisant varier les conditions d'appariement.

En ce qui concerne la tâche 3, nous avons réalisé un système entièrement basé sur des réseaux neuronaux qui combine étiquetage de séquences textuelles et classification pour extraire les informations recherchées des cas cliniques. Si le faible nombre de documents disponibles pour l'apprentissage constituait une contrainte forte pour cette méthode, nous avons choisi d'augmenter le corpus en générant de nouveaux cas cliniques, améliorant ainsi nos résultats. 

Finalement, les résultats obtenus sur les deux tâches semblent encourageants.

\vspace{1.5cm}


\bibliographystyle{taln2019}
\bibliography{deft2019}

\end{document}